\documentclass{article}

\usepackage[final]{corl_2019} % Uncomment for the camera-ready ``final'' version
\usepackage{graphicx} % more modern
\usepackage{amsfonts}
\usepackage{amsmath,soul}
\usepackage[capitalize]{cleveref} %cref
\usepackage{dsfont} % mathds
\usepackage[font=small,skip=0pt]{subcaption} %subfigure
\usepackage{balance}
\usepackage[font=small]{caption}
\usepackage{bm} % bold italic vectors
\usepackage{wrapfig}

\usepackage[markup=nocolor,authormarkuptext=name,addedmarkup=bf,authormarkupposition=left]{changes}
\definechangesauthor[name={J.~H.}, color={blue}]{jh}
\definechangesauthor[name={M.~E.}, color={red}]{me}
\definechangesauthor[name={B.~L.}, color={green}]{bl}
\definechangesauthor[name={G.~H.}, color={orange}]{gh}
\DeclareMathOperator*{\argmin}{\arg\!\min}
\DeclareMathOperator*{\argmax}{\arg\!\max}

\usepackage[sort,compress]{cite}
\usepackage{wrapfig}
\usepackage{tikz}
\usepackage{xcolor}
\colorlet{NextBlue}{red!25!green!50!blue!75}
\usetikzlibrary{shapes,arrows,fit,backgrounds}
\tikzstyle{block} = [draw, rectangle, text width=2cm, text centered, minimum height=1.2cm, node distance=3cm]
\tikzstyle{container} = [draw, rectangle, inner sep=0.3cm, fill=NextBlue!20, minimum height=3cm]
\def\bottom#1#2{\hbox{\vbox to #1{\vfill\hbox{#2}}}}
\usetikzlibrary{positioning,calc}

\tikzset{
  mybackground/.style={execute at end picture={
      \begin{scope}[on background layer]
        \node[font=\bf] at (current bounding box.north){\bottom{1cm} #1};
        \end{scope}
    }},
}

\title{Certified Adversarial Robustness for\\ Deep Reinforcement Learning}
\author{
  Bj\"orn L\"utjens, Michael Everett, Jonathan P. How\\
  Department of Aeronautics and Astronautics \\ Massachusetts Institute of Technology \\
  \texttt{lutjens, mfe, jhow}@mit.edu \\
}

\begin{document}

\maketitle 

%===============================================================================

\begin{abstract}
Deep Neural Network-based systems are now the state-of-the-art in many robotics tasks, but their application in safety-critical domains remains dangerous without formal guarantees on network robustness.
Small perturbations to sensor inputs (from noise or adversarial examples) are often enough to change network-based decisions, which was already shown to cause an autonomous vehicle to swerve into oncoming traffic.
In light of these dangers, numerous algorithms have been developed as defensive mechanisms from these adversarial inputs, some of which provide formal robustness guarantees or certificates.
This work leverages research on certified adversarial robustness to develop an online certified defense for deep reinforcement learning algorithms.
The proposed defense computes guaranteed lower bounds on state-action values during execution to identify and choose the optimal action under a worst-case deviation in input space due to possible adversaries or noise.
The approach is demonstrated on a Deep Q-Network policy and is shown to increase robustness to noise and adversaries in pedestrian collision avoidance scenarios and a classic control task.
\end{abstract}


%===============================================================================

%!TEX root=main.tex

\section{Introduction} \label{sec:intro}
%%Structure:
%neural nets nice, but adversaries dangerous (even for RL and real scenarios)
%especially dangerous for safety-critical tasks, such as pedestrian avoidance
%standard approaches against adversaries come without guarantees
%verification approaches take too long
%certified approaches can be run in real-time and have never been applied to reinforcement learning. 

Deep reinforcement learning (RL) algorithms have achieved impressive success on robotic manipulation~\citep{Gu_2017} and robot navigation in pedestrian crowds~\citep{fan2019getting,Everett_2018}. Many of these systems utilize black-box predictions from deep neural networks (DNNs) to achieve state-of-the-art performance in prediction and planning tasks. However, the lack of formal robustness guarantees for DNNs currently limits their application in safety-critical domains, such as collision avoidance. In particular, even subtle perturbations to the input, known as \textit{adversarial examples}, can lead to incorrect (but highly-confident) predictions from DNNs~\citep{Szegedy_2014, Akhtar_2018, Yuan_2019}.
Furthermore, several recent works have demonstrated the danger of adversarial examples in real-world situations~\citep{Kurakin_2017b, Sharif_2016}, including causing an autonomous vehicle to swerve into oncoming traffic~\citep{Tencent_2019}. The work in this paper addresses the lack of robustness against adversarial examples and sensor noise by proposing an online certified defense to add onto existing deep RL algorithms during execution.

% The reliance of autonomous systems on black-box predictions by neural networks is often inevitable to achieve state-of-the-art performance in prediction or planning tasks. This reliance on neural networks, however, is dangerous for safety-critical tasks, such as collision avoidance. The underlying neural networks are especially vulnerable to subtle perturbations in the input that lead to incorrect and highly-confident predictions, also called adversarial examples~\citep{Szegedy_2014, Akhtar_2018, Yuan_2019}. The vulnerability to adversarial examples is especially critical, as recent works generates adversarial examples in the real-world~\citep{Kurakin_2017b, Sharif_2016} and has, for example, swerved an autonomous vehicle into the oncoming lane~\citep{Tencent_2019}. This work proposes an add-on verification tool for existing Deep RL algorithms to ensure robustness against adversarial examples and sensor noise during test time.

%neural networks did not robustly learn the underlying concepts. 
%face recognition systems have been fooled by adversarial glasses~\citep{Sharif_2016}, and a car has been camouflaged from autonomous vehicle's car detection algorithms~\citep{Zhang_2019}.

Existing methods to defend against adversaries, such as adversarial training~\citep{Kurakin_2017, Madry_2018, Kos_2017}, defensive distillation~\citep{Papernot_2016}, or model ensembles~\citep{Tramer_2018} do not come with theoretical guarantees for reliably improving the robustness and are often ineffective on the advent of more advanced adversarial attacks~\citep{Carlini_2017, He_2017, Athalye_2018, Uesato_2018}.
\textit{Verification} methods do provide formal guarantees on the robustness of a given network, but finding the guarantees is an NP-complete problem and computationally intractable to solve in real-time for applications like robot manipulation or navigation~\citep{Katz_2017, Lomuscio_2017, Tjeng_2019,Ehlers_2017,Huang_2017b}. 
\textit{Robustness certification} methods relax the problem to make it tractable.
Given an adversarial distortion of a nominal input, instead of finding exact bounds on the worst-case output deviation, these methods efficiently find certified lower bounds~\citep{Raghunathan_2018, Wong_2018, Weng_2018, Singh_2018,Wang_2018}.
In particular, the work by~\citep{Weng_2018} runs in real-time for small networks ($33$ to $14,000$ times faster than verification methods), its bound has been shown to be within $10\%$ error of the true bound, and it is compatible with many activation functions and neural network architectures~\citep{Weng_2018b,Boopathy_2019}.
These methods were applied on computer vision tasks.
% Instead of finding exact bounds to the maximum deviation of the output, given an adversarial distortion of the input, it finds certified lower bounds.
% The work by~\cite{Weng_2018}, in particular, achieves a speed-up of $33-14k$ in comparison to verification-based algorithms~\citep{Katz_2017}, can be run in real-time for small networks, delivers a bound usually within $10\%$ error of the true bound, and is extentable to general activation functions~\citep{Weng_2018b} and convolutional neural networks~\citep{Boopathy_2019}. 

\begin{wrapfigure}{r}{0.450\textwidth}
\vspace*{-.15in}
\begin{center}
    \includegraphics[width=0.25\textwidth]{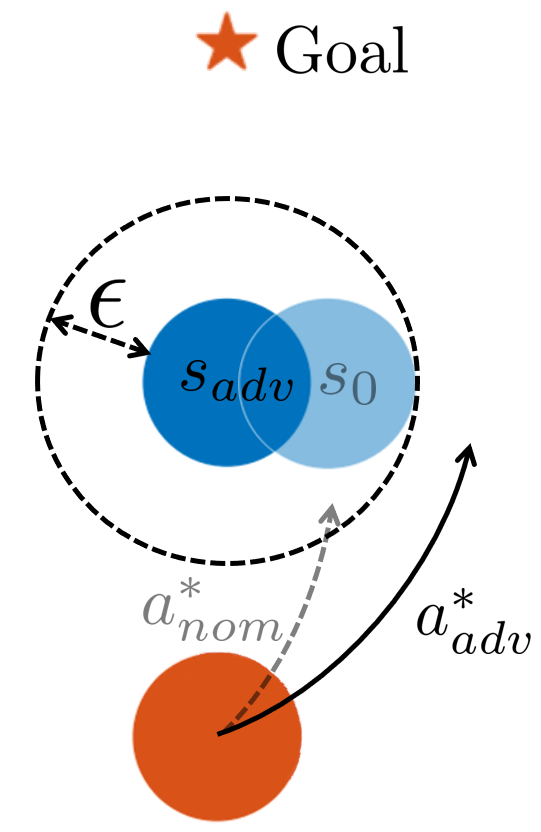}
  \end{center}
  \caption[Intuition on Certified Adversarial Robustness]{Intuition. An adversary distorts the true position, $s_0$, of a dynamic obstacle (blue) into an adversarial observation, $s_{adv}$. The agent (orange) only sees the adversarial input, so nominal RL policies would take $a^*_{nom}$ to reach the goal quickly, but would then collide with the true obstacle, $s_0$. The proposed defensive strategy considers that $s_0$ could be anywhere inside the $\epsilon$-ball around $s_{adv}$, and selects the action, ${a^*_{adv}}$, with the best, worst-case outcome as calculated by a guaranteed lower bound on the value network output, which cautiously avoids the obstacle while reaching the goal. Note this is different from simply inflating the obstacle radius, since the action values contain information about environment dynamics, e.g., blue agent's cooperativeness.}
  \label{fig:intuition_vfy_rl}
\end{wrapfigure}
This work extends the tools for robustness certification against adversaries to deep RL tasks.
As a motivating example, consider the collision avoidance setting in~\cref{fig:intuition_vfy_rl}, in which an adversary perturbs an agent's (orange) observation of an obstacle (blue).
An agent following a nominal/standard deep RL policy would observe $s_{adv}$ and select an action, $a^*_{nom}$, that collides with the obstacle's true position, $s_0$, thinking that space is unoccupied.
Our proposed approach assumes a worst-case deviation of the observed input, $s_{adv}$, bounded by $\epsilon$, and takes the optimal action, $a^*_{adv}$, under that perturbation, to safely avoid the true obstacle.
Nominal robustness certification algorithms assume $\epsilon$ is a scalar, which makes sense for image inputs (all pixels have same scale, e.g., $0{-}255$ intensity).
A key challenge in direct application to RL tasks is that the observation vector (network input) could have elements with substantially different scales (e.g., position, angle, joint torques) and associated measurement uncertainties, motivating our extension with $\epsilon$ as a vector.

This work contributes (i) the first formulation of robustness certification deep RL problems, (ii) an extension of existing robustness certification algorithms to variable scale inputs, (iii) an optimal action selection rule under worst-case state perturbations, and (iv) demonstrations of increased robustness to adversaries and sensor noise on cartpole and a pedestrian collision avoidance simulation.

%!TEX root=main.tex

\section{Related work} \label{sec:related_work}

\subsection{Varieties of adversaries in RL}
RL literature proposes many approaches to achieve adversarial robustness.
Domain Randomization, also called perturbed simulation, adversarially chooses parameters that guide the physics of a simulation, such as mass, center of gravity, or friction during training~\citep{Rajeswaran_2017, Muratore_2018}.
Other work investigates the addition of adversarially acting agents~\citep{Uther_1997, Pinto_2017} during training.
The resulting policies are more robust to a distribution shift in the underlying physics from simulation to real-world, e.g., dynamics/kinematics.
This work, in comparison, addresses adversarial threats in the observation space, not the underlying physics~\citep{Goodfellow_2015}.
For example, adversarial threats could be created by small perturbations in a camera image, lidar pointcloud, or estimated positions/velocities of pedestrians.

% TODO is orthogonal -- hard to understand 
\subsection{Defenses to adversarial examples}
Much of the existing work on robustness against adversarial attacks detects or defends against adversarial examples. %todo_bl reword
Adversarial training or retraining augments the training dataset with adversaries~\citep{Kurakin_2017, Madry_2018, Kos_2017} to increase robustness during testing (empirically). 
Other works increase robustness through distilling networks~\citep{Papernot_2016}, comparing the output of model ensembles~\citep{Tramer_2018}, or detect adversarial examples through comparing the input with a binary filtered transformation of the input~\citep{Xu_2018}.
Although these approaches show impressive empirical success, they do not come with theoretical guarantees for reliably improving the robustness against a variety of adversarial attacks and are often ineffective against more sophisticated adversarial attacks~\citep{Carlini_2017, He_2017, Athalye_2018, Uesato_2018}. 
%~\meXX{what does distill mean}
% ~\meXX{what is the benefit of doing retraining, do you ever find all adverserial points or does this just move them around?}. 

\subsection{Formal robustness verification and certification}
Verification methods provide these desired theoretical guarantees. The methods find theoretically proven bounds on the maximum output deviation, given a bounded input perturbation~\citep{Katz_2017, Lomuscio_2017, Tjeng_2019}. These methods rely on satisfiability modulo theory (SMT)~\citep{Ehlers_2017, Katz_2017, Huang_2017b}, LP, or mixed-integer linear programming (MILP) solvers~\citep{Lomuscio_2017, Tjeng_2019}, or zonotopes~\citep{Gehr_2018}, to propagate constraints on the input space to the output space. The difficulty in this propagation arises through ReLU or other activation functions, which can be nonlinear in between the upper and lower bound of the propagated input perturbation. In fact, the problem of finding the exact verification bounds is NP-complete~\citep{Katz_2017,Weng_2018} and thus currently infeasible to be run online on a robot.
A relaxed version of the verification problem provides certified bounds on the output deviation, given an input perturbation~\citep{Raghunathan_2018, Wong_2018, Weng_2018,Singh_2018,Wang_2018}. Fast-Lin~\citep{Weng_2018}, offers the certification of CIFAR networks in tens of seconds, provable guarantees, and the extendability to all possible activation functions~\citep{Weng_2018b}. This work extends Fast-Lin from computer vision tasks to be applicable in Deep RL domains. 

\subsection{Safe and risk-sensitive reinforcement learning}
Like this work, several Safe RL algorithms surveyed in~\citep{Garcia_2015} also optimize a worst-case criterion. Those algorithms, also called risk-sensitive RL, optimize for the reward under \textit{worst-case} assumptions of environment stochasticity, rather than optimizing for the expected reward~\citep{Heger_1994,Tamar_2015,Geibel_2006}. The resulting policies are more risk-sensitive, i.e., robust to stochastic deviations in the input space, e.g., sensor noise, but could still fail on algorithmically-crafted adversarial examples. To be fully robust against adversaries, this work assumes a worst-case deviation of the input space inside some bounds and takes the action with maximum expected reward. A concurrent line of work trains an agent to not change actions in the presence of adversaries, but does not guarantee that the agent takes the best action under a worst case perturbation~\citep{Fischer_2019}.% This formulation as max-min criterion accounts for the fact that the 
%An agent with this max-min optimization function selects high-reward actions, in comparison, to a concurrent line of work, which trains an agent to not change actions in the presence of adversaries.
%todo_bl take this wording an put into abstract/fig1 
Other work in Safe RL focuses on parameter/model uncertainty, e.g., uncertainty in the model for novel observations (far from training data) in~\citep{Kahn_2017,Lutjens_2019}. Several robotics works avoid the sim-to-real transfer by learning policies online in the real world. However, learning in the real world is slow (requires many samples) and does not come with full safety guarantees~\citep{Sun_2017,Berkenkamp_2017}.

%\subsection{Navigation in Pedestrian Environment}
%A particularly challenging safety-critical task is avoiding pedestrians in a campus environment with an autonomous shuttle bus or rover~\citep{Miller_2016,Navya_2018}. Humans achieve mostly collision-free navigation by understanding the hidden intentions of other pedestrians and vehicles and interacting with them~\citep{Zheng_2015, Helbing_1995}. Furthermore, most of the time this interaction is accomplished without verbal communication. Prior work uses Deep RL to capture the hidden intentions and achieve collaborative navigation around pedestrians~\citep{Chen_2016, Chen_2017, Everett_2018}. This work builds on the formulation of the navigation problem as a partially-observable markov decision process (POMDP).
%\meXX{this seems arbitarily thrown in here. for the thesis the intro probably already motivates this and explains related work?}

\iffalse
PAC-Bayes: Assumes that train and test distribution are the same, I think!

DISCUSS extension to TRPO and A3C
\fi

%!TEX root=main.tex

\section{Background} \label{sec:background}
\subsection{Robustness certification}

In RL problems, the state-action value $Q=\mathbb{E}[\sum_{t=0}^{T}{\gamma^t r_t}]$ expresses the expected future reward, $r_t$, discounted by $\gamma$, from taking an action in a given state/observation. 
This work aims to find the action that maximizes state-action value under a worst-case perturbation of the observation by sensor noise or an adversary.
This section explains how to obtain the certified lower bound on the DNN-predicted $Q$, given a bounded perturbation in the input space from the true state.
The derivation is based on~\citep{Weng_2018}, re-formulated for RL.
We define the certified lower bound of the state-action value, $Q_L$, for each discrete action, $a_j$, as 
\begin{equation}
\begin{aligned}
& Q_L(s_{adv},a_j) := \min_{s \in B_p(s_{adv}, \epsilon)} Q_l(s, a_j), \label{eq:qlj_lower_bnd} \\
\end{aligned}
\end{equation}
for all possible states, $s$, inside the $\epsilon$-Ball around the observed input, $s_{adv} {\in} \mathbb{R}^{n}$, where $B_p(s_{adv}, \epsilon) := \{s : \;\lvert\lvert s - s_{adv} \rvert\rvert_p \leq \epsilon\}$. $Q_l$ is the certified lower bound for a given state: $Q_l(s, a_j)\leq Q(s, a_j) \forall s \in B_p(s_{adv}, \epsilon), \forall a_j{\in}\mathds{A}$, and calculated in~\cref{eq:q_lower_bound}.
The $L_p$-norm bounds the input deviation that the adversary was allowed to apply, and is defined as $\lvert\lvert x \rvert\rvert_p = (\lvert x_1 \lvert^p +  ... + \lvert x_n \lvert^p)^{1/p}$ for $x{\in}\mathbb{R}^{n}$, $p{\geq}1$.
%For example, the $\epsilon$-Ball for $n=2$ and $p=1$ would be a square with the corners on the unit axes, for $p=2$ a unit circle and for $p=\infty$ a square with the edges parallel to the unit axes.
%The particular choices of $p$, $n$, and $\epsilon$ are domain-specific, with suggested choices for two domains in~\cref{sec:results}.
% TODO add dimensionality of s and epsilon

The certification essentially passes the interval bounds $[l^{(0)}, u^{(0)}] = [s_{adv} - \epsilon, s_{adv} + \epsilon]$ from the DNN's input layer to the output layer, where $l^{(k)}$ and $u^{(k)}$ denotes the lower and upper bound of the preReLU-activation, $z^{(k)}$, in the $k$-th layer of an $m$-layer DNN. The difficulty while passing these bounds from layer to layer arises through the nonlinear activation functions, such as ReLU, PReLU, tanh, sigmoid. Note that although this work considers ReLU activations, it can easily be extended to general activation functions via the certification process as seen in~\citep{Weng_2018b}. When passing interval bounds through a ReLU activation, the upper and lower preReLU bound can either both positive $(l^{(k)}, u^{(k)} > 0)$, negative $(l^{(k)}, u^{(k)} < 0)$, or positive and negative $(l^{(k)}<0, u^{(k)}>0)$, in which the ReLU status is called \textit{active, inactive} or \textit{undecided}, respectively. In the active and inactive case, bounds are passed to the next layer as normal. In the undecided case, the output of the ReLU is bounded through linear upper and lower bounds:
\begin{equation}
\begin{aligned}
\sigma_{[l^{(k)}, u^{(k)}]}(z^{(k)})= 
\begin{cases} [z^{(k)}, z^{(k)}] &\text{if $l^{(k)}, u^{(k)} > 0$, ``active''} \\ 
[0,0] &\text{if $l^{(k)}, u^{(k)} < 0$, ``inactive''} \\ 
[\frac{u^{(k)}}{u^{(k)}-l^{(k)}}z^{(k)}, \frac{u^{(k)}}{u^{(k)}-l^{(k)}}(z^{(k)}-l^{(k)})]&\text{if $l^{(k)}{<}0, u^{(k)}{>}0$, ``undecided''.}\end{cases}
\label{eq:sigma_lk_uk}
\end{aligned}
\end{equation}
The identity matrix $D$ is introduced as the ReLU status  matrix, $H$ as the lower/upper bounding factor, $W$ as the weight matrix, $b$ as the bias in layer $(k)$ with $r,j$ as indices, and the preReLU-activation, $z^{(k)}$, is replaced with $W_{r,:}^{(k)}s + b_r^{(k)}$. The ReLU bounding is then rewritten as 
\begin{equation*}
\begin{aligned} 
& D_{r,r}^{(k)}(W_{r,j}^{(k)}s_j + b_r^{(k)}) \leq \sigma(W_{r,j}^{(k)}s_j + b_r^{(k)}) \leq D_{r,r}^{(k)}(W_{r,j}^{(k)}s_j + b_r^{(k)} - H_{r,j}^{(k)}), \\
& \text{where }
D_{r,r}^{(k)} = \begin{cases} \frac{u_r^{(k)}}{u_r^{(k)} - l_r^{(k)}} &\text{if $l_r^{(k)}{<}0, u_r^{(k)}{>}0$;}\\
1 &\text{if $l_r^{(k)}, u_r^{(k)}{>}0$;} \\
0 &\text{if $l_r^{(k)}, u_r^{(k)}{<}0$,} \end{cases} \text{and } 
H_{r,j}^{(k)} = \begin{cases} l_r^{(k)} &\text{if $l_r^{(k)}{<}0, u_r^{(k)}{>}0, A_{j,r}^{(k)}{<}0$;}\\ 0 &\text{otherwise.} \end{cases}
\label{eq:D_H_defns}
\end{aligned}
\end{equation*}
Similar to the closed-form forward pass in a DNN, one can formulate the closed form solution for the guaranteed lower bound of the state-action value for a single state $s$:
\begin{equation}
\begin{aligned}
Q_l(s, a_j) = A_{j,:}^{(0)}s + b_j^{(m)} + \sum_{k=1}^{m-1}{A_{j,:}^{(k)}(b^{(k)}-H_{:,j}^{(k)})},
\label{eq:q_lower_bound}
\end{aligned}
\end{equation}
%TODO: check if this is lower bound inside eps ball or not.
where the matrix $A$ contains the network weights and ReLU activation, recursively for all layers: $A^{(k-1)}=A^{(k)}W^{(k)}D^{(k-1)}$, with identity in the final layer: $A^{(m)}=\mathds{1}$.

\subsection{Pedestrian simulation}\label{sec:approach:pedestrian_simulation}
Among the many RL tasks, a particularly challenging safety-critical task is collision avoidance for a robotic vehicle among pedestrians.
Because learning a policy in the real world is dangerous and time consuming, this work uses a kinematic simulation environment for learning pedestrian avoidance policies.
The decision process of an RL agent in the environment can be described as Partially Observable Markov Decision Process (POMDP) with the tuple $<S, \mathds{A}, T, R, \Omega, O, \gamma>$.
The environment state $S$ is fully described by the behavior policy, position, velocity, radius, and goal position of each agent.
In this example, the RL policy controls one of two agents with 11 discrete action heading actions $A = [a_{\min}, a_{\max}] = [-\pi/6, +\pi/6]$ and constant velocity $v=1m/s$.
The environment executes the selected action under unicycle kinematics, and controls the other agent from a diverse set of fixed policies (static, non-cooperative, ORCA~\citep{Berg_2009}, GA3C-CADRL~\citep{Everett_2018}).
The sparse reward is $1$ for reaching the goal, $-0.25$ for colliding and the partial observation is the x-y position, x-y velocity, and radius of each agent, and the RL agent's goal, as in~\citep{Everett_2018}.
%!TEX root=main.tex

\section{Approach} \label{sec:approach}
This work develops an add-on certified defense for existing Deep RL algorithms to ensure robustness against sensor noise or adversarial examples during test time.

\subsection{System architecture}

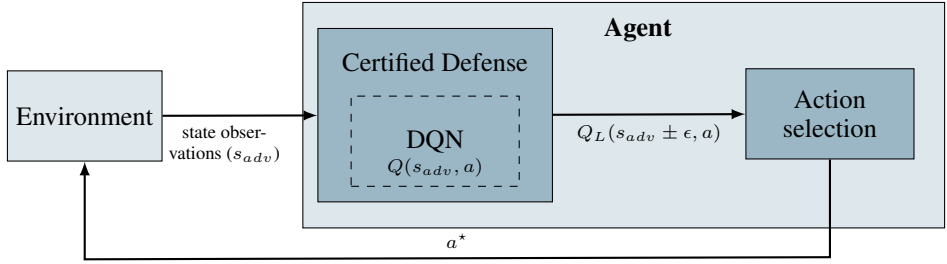
\begin{figure}[t]
  \centering
    \centering

\begin{tikzpicture}[]%mybackground={Agent}

    \node [draw, text centered, rectangle, minimum height=1.2cm, fill=NextBlue!20, name=text1] {Environment};
    \node [block, fill=NextBlue!60, above right=-1.55cm and 2.5cm of text1, dashed] (text2) {DQN};
    %\node [block, fill=NextBlue!60, right of=text2] (text3) {Certification};
    \node (text7) [below =-.5cm of  text2, font=\scriptsize]  {$Q(s_{adv}, a)$};
    \node (text5) [above =0.2cm of  text2]  {Certified Defense}; 
    \node [block, fill=NextBlue!60, above right=-.85cm and 3.0cm of text2] (text4) {Action\\ selection};
%    \node [block, right of=text4] (text5) {Text5};
    \begin{scope}[on background layer]
    \node [container,fit=(text2) (text5) (text4),inner sep=0.4cm] (container) {};
    \end{scope}
    \begin{scope}[on background layer]
    \node[fit= (text5) (text2),draw,fill=NextBlue!60, inner sep=0.2cm] (Box)   {};
    \end{scope}
    \node (text6) [above right=0.6cm and 1.0cm of text2]  {\textbf{Agent}};
    \draw [-latex,thick] (text1) -- (Box) node[align=center,draw=none,fill=none,font=\scriptsize,pos=0.4,below]{state obser-\\vations ($s_{adv}$)};
    % \draw [-latex,thick] (text2) -- (text3) node[align=center,draw=none,fill=none,font=\scriptsize,midway,above]{W};
    \draw [-latex,thick] (Box) -- (text4) 
 node[align=center,draw=none,fill=none,font=\scriptsize,midway,below]{$Q_L(s_{adv}\pm\epsilon,a)$};
%    \draw [->] () |- node {} (text1.south);
  \draw [-latex,thick] (text4.south) --  ($ (text4.south) - (0cm,1.3cm) $) -- node[align=center,draw=none,fill=none,font=\scriptsize,midway,above] {$a^\star$} ($ (text1.south) - (0cm,1.3cm) $) -- ($ (text1.south) - (0,0) $); %
    %\draw [-latex, thick] ($(text2.west) - (.175cm,0) $) |- ($(text3.north) + (0,.275cm) $) node[align=center,draw=none,fill=none,font=\scriptsize,right,above]{$s\pm\epsilon$} -- ($(text3.north) - (0,0) $);
  %\draw [-latex,thick] ($(text2.south) - (0,0cm) $) |-  ($(text3.south) - (0,.5cm) $) node[align=left ,draw=none, fill=none, font=\scriptsize, pos=0.8, above]{$Q(s,a)$} -- ($(text3.south) - (0,0) $);
  \end{tikzpicture}

\caption[System Architecture for Certified Adversarial Robustness for Deep Reinforcement Learning]{System Architecture. During online execution, an agent observes a state, $s_{adv}$, corrupted by sensor noise or an adversarial attack. A Deep RL algorithm, e.g., Deep Q-Network (DQN)~\citep{Mnih_2015}, predicts the state-action values, $Q$. A node for certified defense accesses the predicting network, adds a robustness threshold $\pm\epsilon$ in the input space and computes a lower bound of the state-action values of each discrete action: $Q_L$. The agent takes the action, $a^*$, that maximizes the lower bound, i.e. is the most robust to the deviation in the input space.}
  \label{fig:sys_arch} 
  \vspace{-.2in}
\end{figure}

\Cref{fig:sys_arch} depicts the system architecture of a standard model-free RL framework with the added-on certification.
In an offline training phase, an agent uses a deep RL algorithm, here DQN~\citep{Mnih_2015}, to train a DNN that maps non-corrupted state observations, $s$, to state-action values, $Q(s,a)$.
Action selection during training uses the nominal cost function, $a^*_{nom} = \argmax_a Q(s, a)$.
We assume the training process causes the network to converge to the optimal value function, $Q^*(s,a)$ and focus on the challenge of handling perturbed observations during execution.%todo_bl I guess, I could have trained with full and not partial observability to illustrate the paper better

During online execution, the agent only receives corrupted state observations from the environment, and passes those through the DNN.
The certification node uses the DNN architecture and weights, $W$, to compute lower bounds on $Q$ under a bounded perturbation of the input $s\pm\epsilon$, which are used for robust action selection during execution (described below).

\subsection{Optimal cost function under worst-case perturbation}
We consider robustness to an adversary who picks the worst-possible state observation, $s_{adv}$, within a small perturbation, $\epsilon$, of the true state, $s_0$.
The adversary assumes the RL agent follows a nominal policy (as in e.g., DQN) of selecting the action with highest Q-value at the current observation.
% Without consideration of adversaries or noise, standard DQN follows the standard optimal action $a^*_{std}$ that maximizes the expected reward, given an observation of state $s$: $a^*_{std} = \argmax_a Q(s,a)$.
A worst possible state observation, $s_{adv}$, is therefore any one which causes the RL agent to take the action with lowest Q-value in the true state $s_0$,
% todo_bl: make clear why choosing this a* makes sense... {Knowing this, the worst possible state observation, $s_{adv}$, would be the one that minimizes the expected reward of the standard optimal action}
\begin{equation}
s_{adv} \in \{s:\; s \in B_p(s_0, \epsilon)\, \text{and}\, \argmax_a Q(s, a) = \argmin_a Q(s_0, a)\}
%\argmin_{s\in B_p(s_0, \epsilon)}\max_a Q(s,a).\\
\label{eq:s_adv} 
\end{equation}

After the adversary picks the state observation, the agent selects an action.
Instead of trusting the observation (and thus choosing the worst action for the true state), the agent leverages the fact that the true state $s_0$ could be anywhere inside an $\epsilon$-Ball around $s_{adv}$: $s_0 \in B_p(s_{adv}, \epsilon)$.
The agent evaluates each action by calculating the worst-case Q-value under all the possible true states.
The optimal action, $a^*$, is defined here as one with the highest Q-value under the worst-case perturbation,
\begin{equation}
 a^* = \argmax_{a_j}\min_{s\in B_p(s_{adv}, \epsilon)} Q_l(s,a_j) = \argmax_{a_j} Q_L(s_{adv}, a_j),
\label{eq:opt_cost_fn} 
\end{equation}
using the outcome of the certification process, $Q_L$, as defined in~\cref{eq:qlj_lower_bnd}.

% The worst possible action would be the standard optimal action $a_{adv} = \argmax_a Q(s_{adv}, a)$ and all other actions would be better or equal to $a_{adv}$.

% To find the best out of the other possible action in a safety-critical domain, we consider acting conservatively.
% The conservative best action has to assume that the true state could be anywhere inside an $\epsilon$-Ball around $s_{adv}$.
% The agent evaluates each action by estimating the worst-case outcome under all the possible true states.
	
\subsection{Adapting robustness certification to deep RL}
To solve~\cref{eq:opt_cost_fn} when $Q(s,a)$ is represented by a DNN, we adapt the formulation from~\citep{Weng_2018}.
%The goal of~\citep{Weng_2018} is to certify the robustness on a classification task, i.e. provide the minimum adversarial distortion, that could cause a misclassification.
%We use a \jhmargin{pre-computation to adapt}{not sure what this means} the tool to robustness on a regression task, i.e. certifying that an output will stay within some bounds, given an adversarial distortion.~\meXX{agreed - delete or re-word.}
Most works in adversarial examples, including~\citep{Weng_2018}, focus on defending adversaries on image inputs, in which all channels have the same scale, e.g. black/white images with intensities in $[0,255]$.
More generally, however, input channels could be on different scales, e.g. joint torques, velocities, positions.
Although not often mentioned, these sensor readings can also be prone to adversarial attacks, if transmitted over an insecure messaging framework, e.g. ROS, or accidentally producing adversaries through sensor failure or noise.
Hence, this work extends~\citep{Weng_2018} to certify robustness of variable scale inputs and enables the usage to the broader robotics and machine learning community.

To do so, we compute the lower bound $Q_L(s_{adv}, a_j)$ for all states inside the $\epsilon$-Ball $B_p(s_{adv},\epsilon)$ around $s_{adv}$ similar to~\citep{Weng_2018}, but with vector $\epsilon$ (instead of scalar $\epsilon$): % TODO_bl: explain what's in the \epsilon vector
%Reasonal behind minimax Q-Values -- proof on Dynamic Programming Q-learning?! -- Principle of Optimality
%Argue extensible to A3C, TRPO, DDPG (continuous action space)
%Argue extension to GA3C-CADRL
\allowdisplaybreaks
\begin{align}
 Q_L(s_{adv}, a_j) &= \min_{s\in B_p(s_{adv},\epsilon)}\left(A_{j,:}^{(0)}s + b_j^{(m)} + \sum_{k=1}^{m-1}{A_{j,:}^{(k)}(b^{(k)}-H_{:,j}^{(k)})}\right)\label{eq:qlj_full}\\
& = \left(\min_{s\in B_p(s_{adv},\epsilon)}A_{j,:}^{(0)}s\right) + b_j^{(m)} + \sum_{k=1}^{m-1}{A_{j,:}^{(k)}(b^{(k)}-H_{:,j}^{(k)})}\label{eq:qlj_s}\\
& = \left(\min_{y\in B_p(0, 1)}A_{j,:}^{(0)} (y\circ\epsilon)\right) + A_{j,:}^{(0)}s_{adv} + b_j^{(m)} + \sum_{k=1}^{m-1}{A_{j,:}^{(k)}(b^{(k)}-H_{:,j}^{(k)})}\label{eq:qlj_y}\\ 
& = \left(\min_{y\in B_p(0, 1)}(\epsilon\circ A_{j,:}^{(0)}) y\right) + A_{j,:}^{(0)}s_{adv}  + b_j^{(m)} + \sum_{k=1}^{m-1}{A_{j,:}^{(k)}(b^{(k)}-H_{:,j}^{(k)})}\label{eq:qlj_shuffle}\\ 
& = -\lvert\lvert\epsilon\circ A_{j,:}^{(0)}\rvert\rvert_q + A_{j,:}^{(0)}s_{adv}  + b_j^{(m)} + \sum_{k=1}^{m-1}{A_{j,:}^{(k)}(b^{(k)}-H_{:,j}^{(k)})}\label{eq:qlj_norm},
\end{align}
% \label{eq:var_scale_eps} 
with $\circ$ denoting element-wise multiplication.
From~\cref{eq:qlj_s} to~\cref{eq:qlj_y}, substitute ${s := y \circ \epsilon + s_{adv}}$, to shift and re-scale the observation to within the unit ball around zero, $y\in B_p(0,1)$.
The maximization in~\cref{eq:qlj_shuffle} reduces to a q-norm in~\cref{eq:qlj_norm} by the definition of the dual norm $\lvert\lvert z \rvert\rvert_* = \{\text{sup} z^Ty:\lvert\lvert y \rvert\rvert \leq 1\}$ and the fact the $l_q$ norm is dual of $l_p$ norm for $p,q \in [1,\infty)$ (with $1/p + 1/q = 1$). The closed form in~\cref{eq:qlj_norm} is inserted into~\cref{eq:opt_cost_fn} to return the best action.

% \iffalse
% However, adversarial examples are defined as small perturbations in the input space and noise in the observations is also, typically, close to the input space. Hence, the true input must lie inside a ball of small perturbations of the observed adversarial input: $\bm s \pm \bm \epsilon$. $\epsilon$ is a variable scale vector of considered deviations in the input space, e.g. $\epsilon = [\epsilon_{pos_x}, \epsilon_{pos_y}, \epsilon_{vel_x}, \epsilon_{vel_y}] = [0.2m, 0.2m, 0.1\frac{m}{s}, 0.1\frac{m}{s}]$. Finally, this framework takes the most robust action, i.e. the action that maximizes reward, under consideration of the worst possible input: $a* = \argmax_a \min_{s_\epsilon} Q(s_\epsilon, a)$, where $\{s_\epsilon \lvert s \in \bm s\pm\bm \epsilon\}$.

% \fi

%!TEX root=main.tex

\section{Experimental Results} \label{sec:results}
The robustness against adversaries and noise during execution is evaluated in simulations for collision avoidance among pedestrians and the cartpole domain.
In both domains, increasing magnitudes of noise or adversarial attacks are added onto the observations, which reduces the reward of an agent following a nominal non-robust DQN policy.
The added-on defense with robustness parameter, $\epsilon$, increases the robustness of the policy to the introduced perturbations and increases the performance.

\subsection{Adversarially robust collision avoidance}
A nominal DQN policy was trained in the environment described in~\cref{sec:approach:pedestrian_simulation}.
To evaluate the learned policy's robustness to deviations of the input in an $\epsilon$-ball, $B_{p}(s_0,\epsilon)$, around the true state, $s_0$, the observations of the environment agent's position are deviated by an added uniform noise ${\sim}\mathbb U([-\sigma, \sigma])$, or adversarial attack during testing. The adversarial attack is a fast gradient sign method with target (FGST)~\citep{Kurakin_2017b} and approximates the adversary from~\cref{eq:s_adv}. FGST crafts the state $\hat s_{adv}$ on the $\epsilon$-Ball's perimeter that maximizes the Q-value for the nominally worst action $\argmin_a Q(s_0, a)$. Specifically, $\hat s_{adv}$ is picked along the direction of lowest softmax-cross-entropy loss, $\mathbb{L}$, in between a one-hot encoding $y_{adv}$ of the worst action and the nominal Q-values, $y_{nom}$:
\begin{equation*}
\begin{aligned}
& \hat s_{adv} = s_0 - \epsilon_{adv}\; \text{sign}(\nabla_s \mathbb{L}(y_{adv}, y_{nom}))\\
& y_{adv} = [\mathds{1} \{a_i = {\mathop{\mathrm{argmin}}\nolimits}_a Q(s_0, a)\}]  \in \mathbb{R}^{\lvert \mathds{A} \rvert} \\
& y_{nom} = [Q(s_0, a_i)] \in \mathbb{R}^{\lvert \mathds{A} \rvert}
\end{aligned}
\label{eq:s_adv_hat} 
\end{equation*}

\begin{figure}%[tp]
	\centering
	\begin{minipage}[t]{0.45\linewidth}
	\centering\includegraphics [trim=0 0 0 30, clip, width=1.0\textwidth, angle = 0]{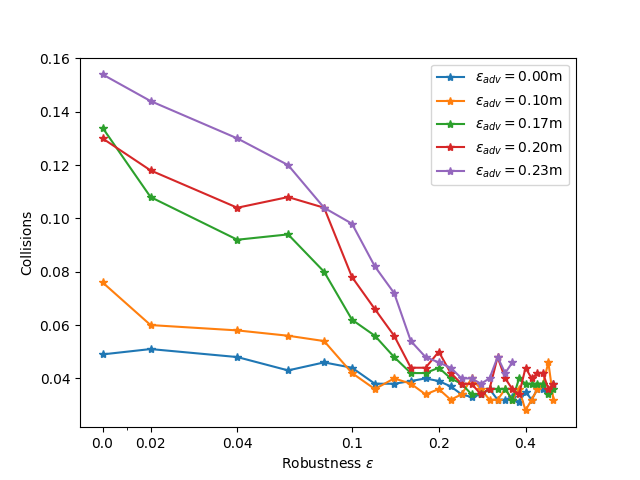}
	\vspace{-0.15in}
	\subcaption{}\label{fig:2d_adv_colls}
	\end{minipage}%
	\begin{minipage}[t]{0.45\linewidth}
	\centering\includegraphics [trim=0 0 0 30, clip, width=1.0\textwidth, angle = 0]{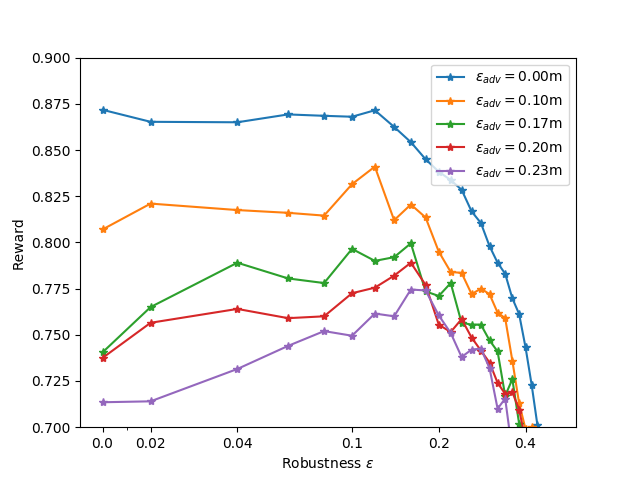}
	\vspace{-0.15in}
	\subcaption{}\label{fig:2d_adv_rew}
	\end{minipage}\\
	\begin{minipage}[t]{0.45\linewidth}
	\centering\includegraphics [trim=0 0 0 30, clip, width=1.0\textwidth, angle = 0]{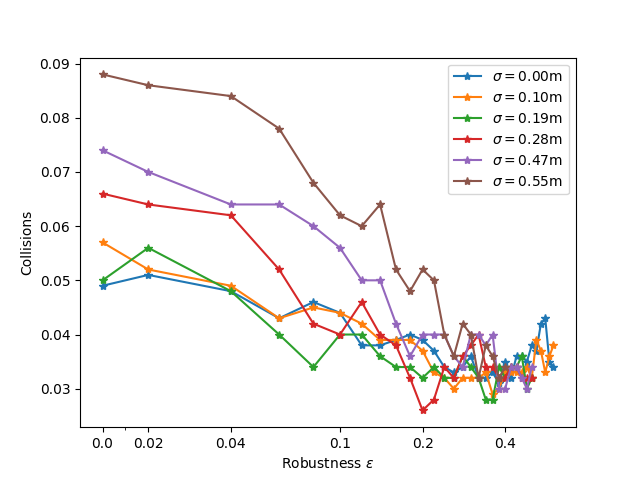}
	\vspace{-0.15in}
	\subcaption{}\label{fig:2d_colls}
	\end{minipage}%
	\begin{minipage}[t]{0.45\linewidth}
	\centering\includegraphics [trim=0 0 0 30, clip, width=1.0\textwidth, angle = 0]{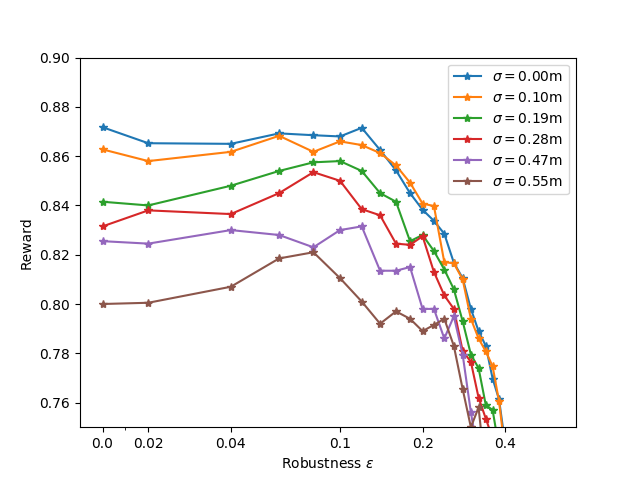}
	\vspace{-0.15in}
	\subcaption{}\label{fig:2d_rew}
	\end{minipage}
	\vspace{-.05in}
\caption[Robustness against adversaries]{Robustness against adversarial attacks and noise.~\Cref{fig:2d_adv_colls,fig:2d_colls} shows that an increasing robustness parameter, $\epsilon$, decreases the number of collisions in the existence of adversarial attacks, or noise with increasing magnitude $\epsilon_{adv}, \sigma$.~\Cref{fig:2d_adv_rew,fig:2d_rew} show that an increasing robustness parameter $\epsilon$ increases the reward for several magnitudes of adversarial attacks or noise (e.g., $\epsilon_{adv}{=}0.17m, \sigma{=}0.28m$) while $\epsilon < {\sim0.2}$.}\label{fig:robustness_eps_adv}
\vspace{-.1in}
\end{figure}

As expected, the nominal DQN policy, $\epsilon{=}0$, is not robust to the perturbation of inputs: Increasing the magnitude of adversarial, or noisy perturbation, $\epsilon_{adv}, \sigma$ drastically 1) increases the average number of collisions (as seen in ~\cref{fig:2d_colls,fig:2d_adv_colls}, respectively, at $\epsilon{=}0$) and 2) decreases the average reward (as seen in ~\cref{fig:2d_rew,fig:2d_adv_rew}). The number of collisions and the reward are reported per run and have been averaged over $5$x$100$ episodes in the stochastic environment. Every set of $500$ episodes constitutes one data point, $*$, and has been initialized with the same $5$ random seeds.
%
%1) The average number of collisions is drastically increased with an increasing magnitude of perturbation $\sigma, \epsilon_{adv}$, as seen in~\cref{fig:2d_colls,fig:2d_adv_colls} (at $\epsilon{=}0$). 2) 

%the added noise, or the adversarial attacks, as seen in~\cref{fig:robustness_eps_adv}. The average number of collisions drastically increases with an increasing perturbation magnitude $\sigma, \epsilon_{adv}$, as seen in~\cref{fig:2d_colls,fig:2d_adv_colls}, respectively (at $\epsilon{=}0$). Consecutively, the average reward decreases with an increasing perturbation magnitude $\sigma, \epsilon_{adv}$, as seen in~\cref{fig:2d_rew,fig:2d_adv_rew}. 

Next, we demonstrate the effectiveness of the proposed add-on defense, called Certified Adversarially-Robust Reinforcement Learning (CARRL). 
The number of collisions decreases with an increasing robustness parameter $\epsilon$ under varying magnitudes of noise, or adversarial attack, as seen in~\cref{fig:2d_adv_colls,fig:2d_colls}. 
As a result, the reward increases with an increasing robustness parameter $\epsilon{<}{\sim}0.1$ under varying magnitudes of perturbations, as seen in~\cref{fig:2d_adv_colls,fig:2d_colls}.
As expected, the importance of the proposed defense is highest under strong perturbations, as seen in the strong slopes of the curves $\epsilon_{adv}{=}0.23m$ and $\sigma{=}0.55m$. Interestingly, the CARRL policy only marginally reduces the reward under no perturbations, $\sigma{=}0, \epsilon_{adv}{=}0$.% This suggests that CARRL could be run on a robot even if there would just be a low probability for sensor noise or adversarial attacks.

\begin{figure}%[tp]
	\vspace{-0.15in}
	\centering
	\begin{minipage}[b]{0.16\linewidth}
	\centering\includegraphics [trim=0 0 0 80, clip, width=1.0\textwidth, angle = 0]{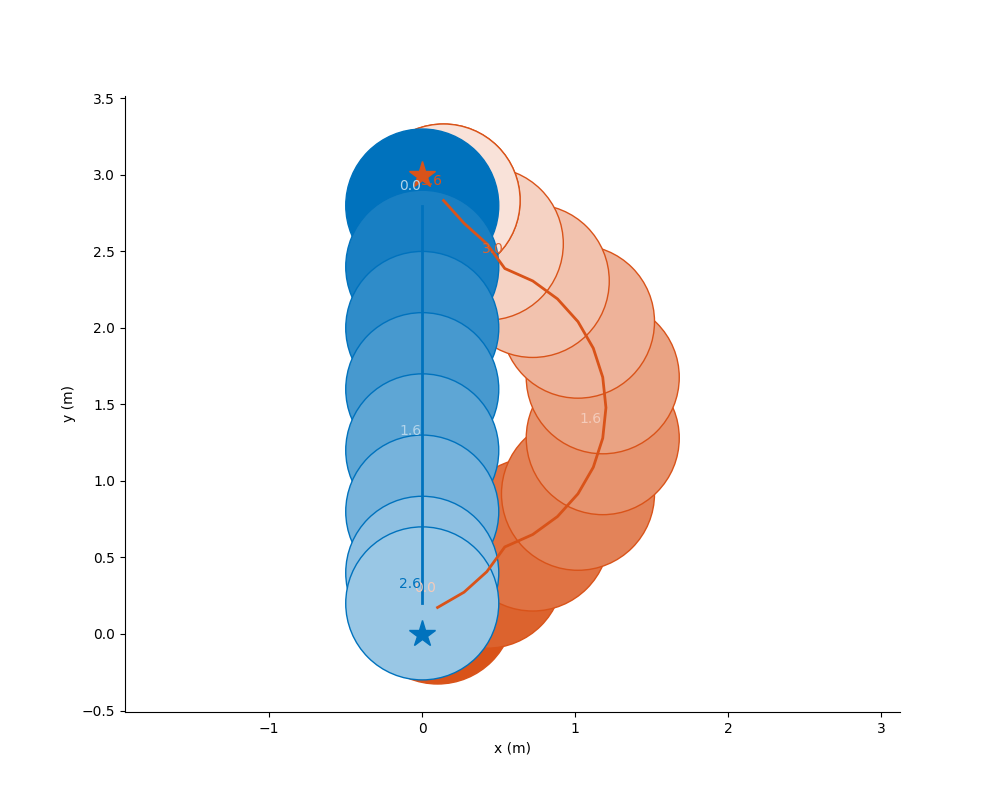}
	\vspace{-0.15in}
	\subcaption{$\epsilon=0.0$}\label{fig:eps_visual1}
	\end{minipage}%
	\begin{minipage}[b]{0.16\linewidth}
	\centering\includegraphics [trim=0 0 0 80, clip, width=1.0\textwidth, angle = 0]{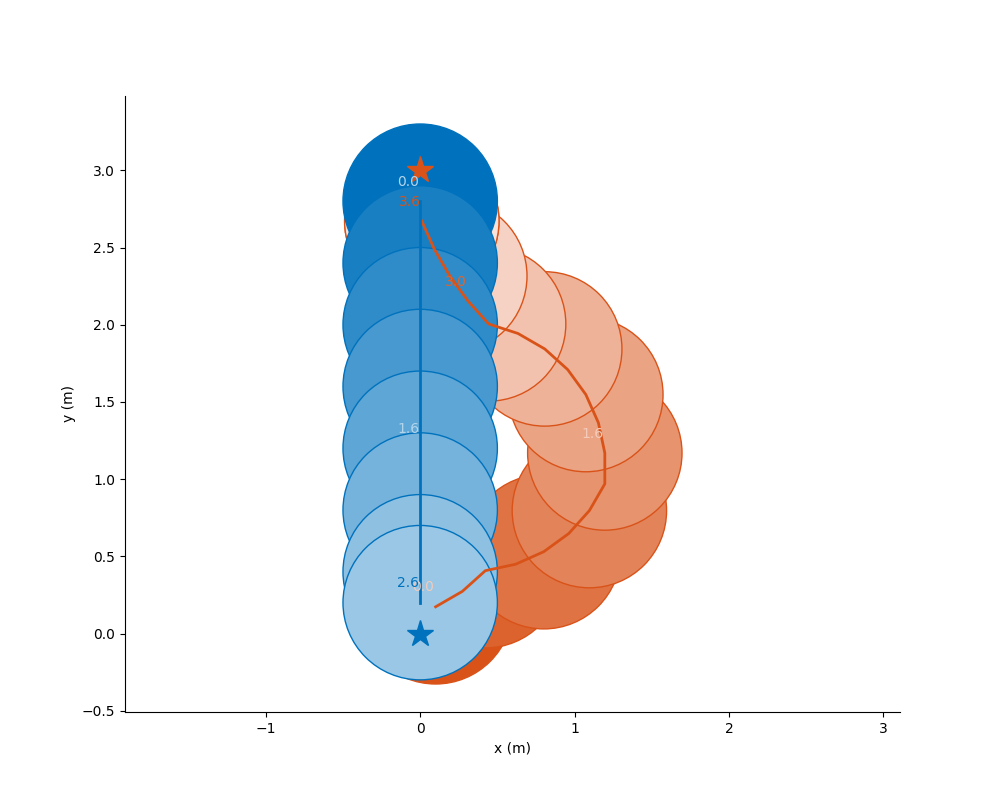}
	\vspace{-0.15in}
	\subcaption{$\epsilon=0.1$}\label{fig:eps_visual2}
	\end{minipage}
	\begin{minipage}[b]{0.16\linewidth}
	\centering\includegraphics [trim=0 0 0 80, clip, width=1.0\textwidth, angle = 0]{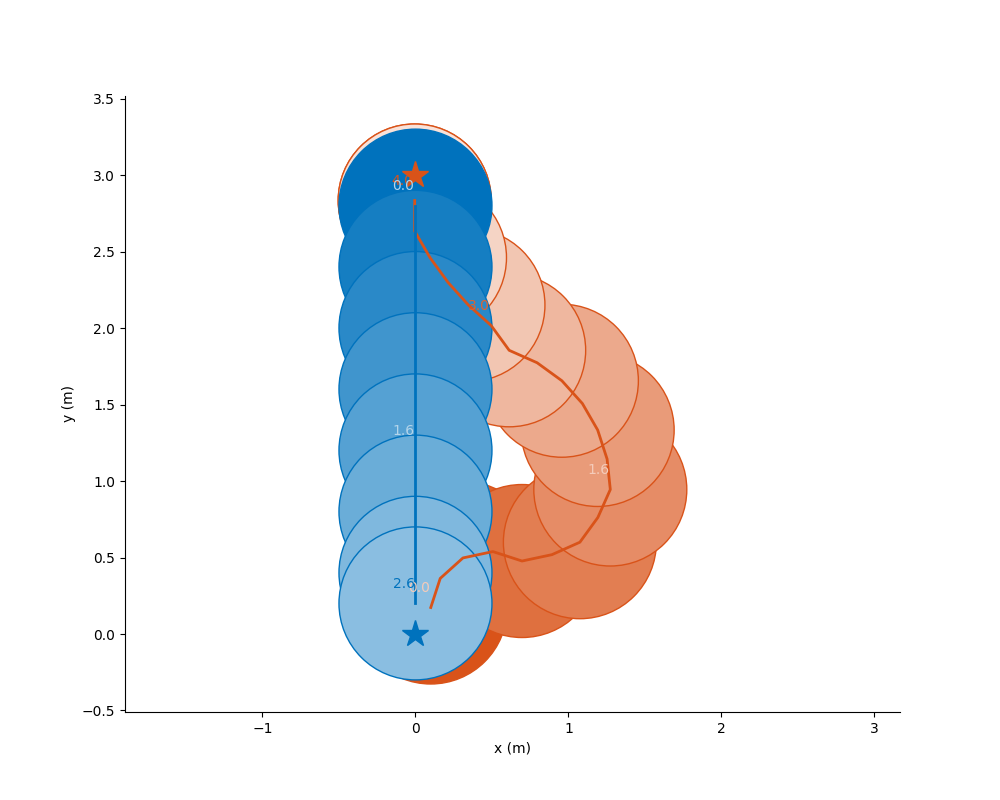}
	\vspace{-0.15in}
	\subcaption{$\epsilon=0.2$}\label{fig:eps_visual3}
	\end{minipage}%\\
	\begin{minipage}[b]{0.16\linewidth}
	\centering\includegraphics [trim=0 0 0 80, clip, width=1.0\textwidth, angle = 0]{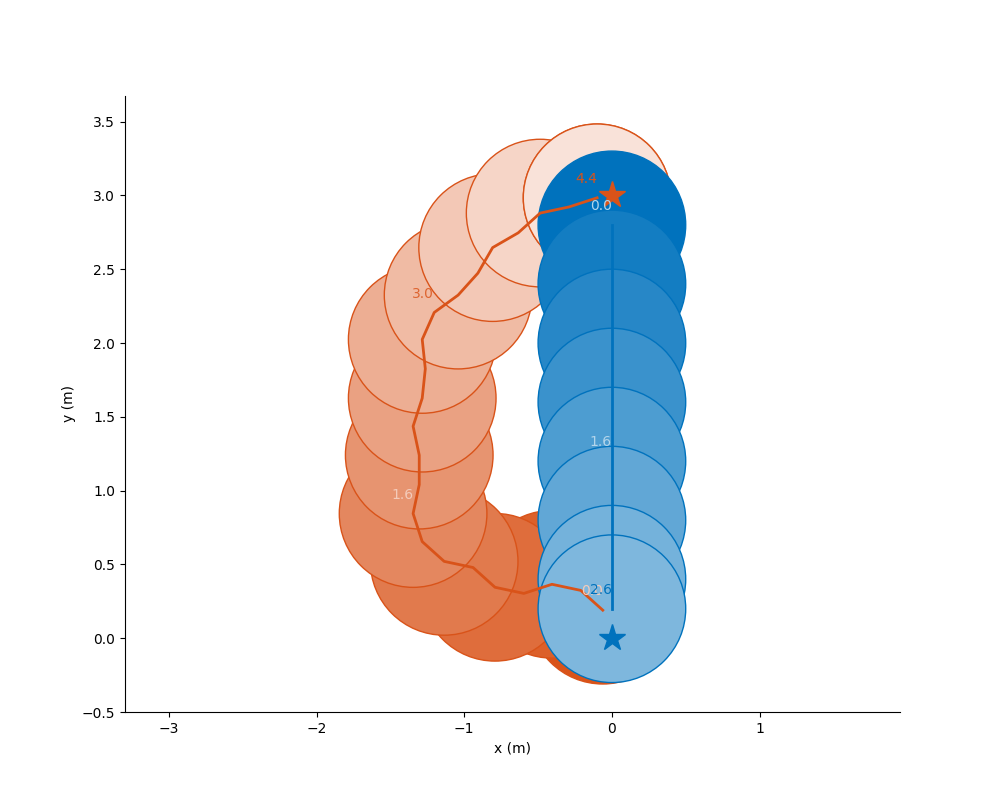}
	\vspace{-0.15in}
	\subcaption{$\epsilon=0.3$}\label{fig:eps_visual4}
	\end{minipage}%
	\begin{minipage}[b]{0.16\linewidth}
	\centering\includegraphics [trim=0 0 0 80, clip, width=1.0\textwidth, angle = 0]{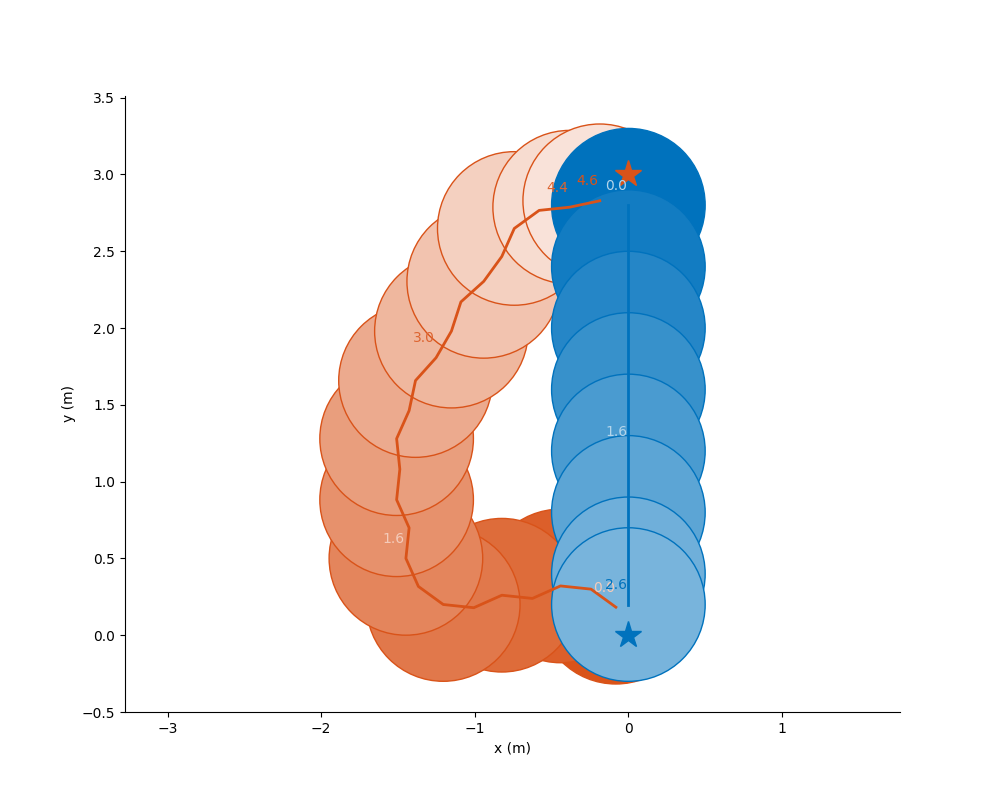}
	\vspace{-0.15in}
	\subcaption{$\epsilon=0.4$}\label{fig:eps_visual5}
	\end{minipage}
	\begin{minipage}[b]{0.16\linewidth}
	\centering\includegraphics [trim=0 0 0 80, clip, width=1.0\textwidth, angle = 0]{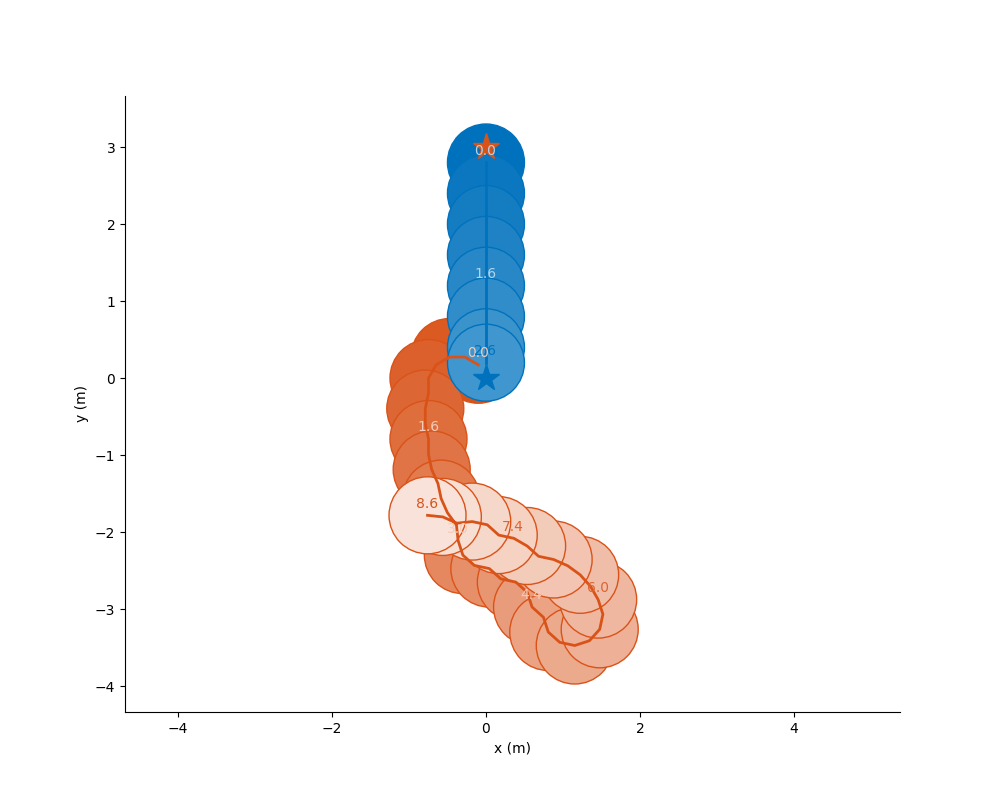}
	\vspace{-0.15in}
	\subcaption{$\epsilon=1.0$}\label{fig:eps_visual6}
	\end{minipage}
\caption[Increase of conservatism with increased $\epsilon$ robustness]{Increase of conservatism with $\epsilon$. An agent (orange) following the CARRL policy avoids a dynamic, non-cooperative obstacle (blue) that is observed without noise. An increasing robustness parameter $\epsilon$ (left to right) increases the agent's conservatism, i.e., the agent avoids the obstacle with a greater safety distance.}
\label{fig:eps_visual}
\vspace{-0.2in}
\end{figure}

Since the CARRL agent selects actions more conservatively than a nominal DQN agent, it is able  successfully reach its goal instead of colliding like the nominal DQN agent does under noisy or adversarial perturbations.
Interestingly, the reward drops significantly for $\epsilon{>}{\sim}0.1$, because the agent ``runs away'' from the obstacle and never reaches the goal, also seen in~\cref{fig:eps_visual6}.
This is likely due to the relatively small exploration of the full state space while training the Q-network.
$\epsilon{>}0.1$ yields an $\epsilon$-ball around $s_{adv}$ that is too large and contains states that are far from the training data, causing our learned Q-function to be inaccurate, which breaks CARRL's assumption of a perfectly learned Q-function for $\epsilon>{\sim}0.1$.
However, even with a perfectly learned Q-function, the agent could run away or stop, when all possible actions could lead to a collision, similar to the \textit{freezing robot problem}~\citep{Trautman2010}.

\begin{figure}%[tp]
	\vspace{-0.1in}
	\centering
	\begin{minipage}[t]{0.35\linewidth}
	\centering\includegraphics [trim=0 0 0 20, clip, width=1.0\textwidth, angle = 0]{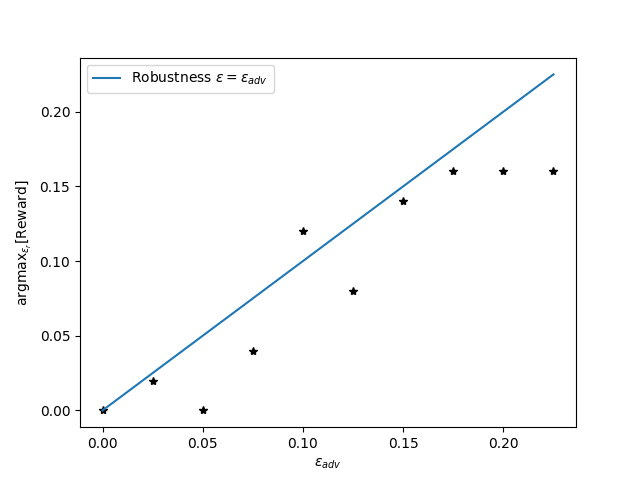}
	\vspace{-0.15in}
	\subcaption{}\label{fig:2d_adv_atk_corr}
	\end{minipage}%
	\begin{minipage}[t]{0.35\linewidth}
	\centering\includegraphics [trim=0 0 0 20, clip, width=1.0\textwidth, angle = 0]{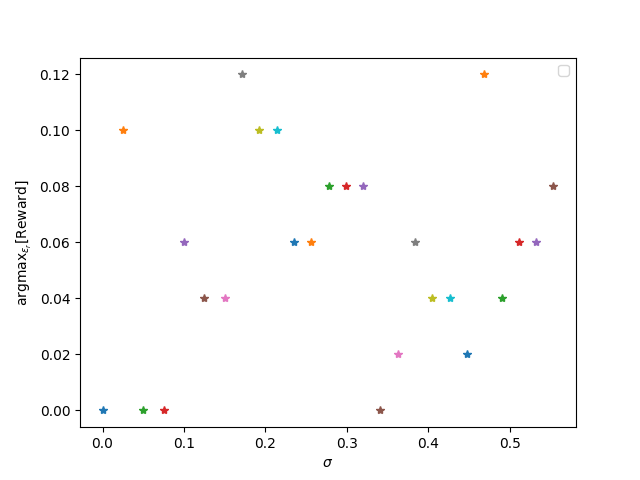}
	\vspace{-0.15in}
	\subcaption{}\label{fig:2d_noise_corr}
	\end{minipage}
    \vspace{-.05in}
\caption[Correlation]{Correlation between perturbation magnitude and $\epsilon$ robustness.~\Cref{fig:2d_adv_atk_corr} shows that the magnitude of the adversarial attack, $\epsilon_{adv}$, is linearly correlated with the best robustness, $\epsilon$ (i.e., one that maximizes the reward under a given attack magnitude).~\Cref{fig:2d_noise_corr} shows that this correlation does not exist when defending against noise. A possible explanation is that an adversary chooses an input state on the perimeter of the $\epsilon$-ball, whereas uniform noise samples from inside the $\epsilon$-ball.}\label{fig:corr_eps_adv}
\vspace{-.25in}
\end{figure}

\Cref{fig:corr_eps_adv} illustrates further intuition on choosing $\epsilon$.~\Cref{fig:2d_adv_atk_corr} demonstrates a linear correlation in between the attack magnitude $\epsilon_{adv}$ and the best defending $\epsilon$ from~\cref{fig:2d_adv_rew}, i.e., the $\epsilon$ that maximizes the reward under the given attack magnitude $\epsilon_{best}=\argmax_{\epsilon \in [\epsilon_{min}, \epsilon_{max}]}R(\epsilon, \epsilon_{adv})$. A correlation cannot be observed under sensor noise, as seen in~\cref{fig:2d_noise_corr}, because an adversary chooses an input state on the perimeter of the $\epsilon$-ball, whereas uniform noise samples from inside the $\epsilon$-ball.
The $p{=}\infty$-norm has been chosen for robustness against uniform noise and the FGST attack, as position observations could be anywhere within a 2-D box around the true state. The modularity in $\epsilon$ allows CARRL to capture uncertainties of varying scales in the input space, e.g., $\epsilon$ is here non-zero for position and zero for other inputs, but could additionally be set non-zero for velocities. $\epsilon$ can further be adapted on-line to account for, e.g., time-varying sensor noise.
%This reduction in collisions comes from the fact that CARRL assumes the worst-case deviation of the observed state inside the $B_{p}(s_{adv}, \epsilon)$-Ball.

The CARRL policy is able run at real-time; querying the policy took ${\sim}20ms$. One forward pass with certified bounds took ${\sim}2ms$, which compares to a forward pass of our nominal DQN of ${\sim}1ms$. In our implementation, we inefficiently query the network once for each of the $11$ actions. The runtime could be reduced by parallelizing the action queries and should remain low for larger networks, as ~\citep{Weng_2018} shows that the runtime scales linearly with the network size.
%TODO The best $\epsilon$ depends on the noise during test time. One would expect that the best $\epsilon$ equals the uniform noise. However, nominal DQN is already robust to some degree of noise and $\epsilon$ is more like the add-on robustness. Nevertheless, the reward maximizing $\epsilon$ increases with noise as depicted in~\cref{fig:3d_rew}. 

\begin{figure}%[tp]
	\centering
	\begin{minipage}[t]{0.35\linewidth}
	\centering\includegraphics [trim=0 0 0 80, clip, width=1.0\textwidth, angle = 0]{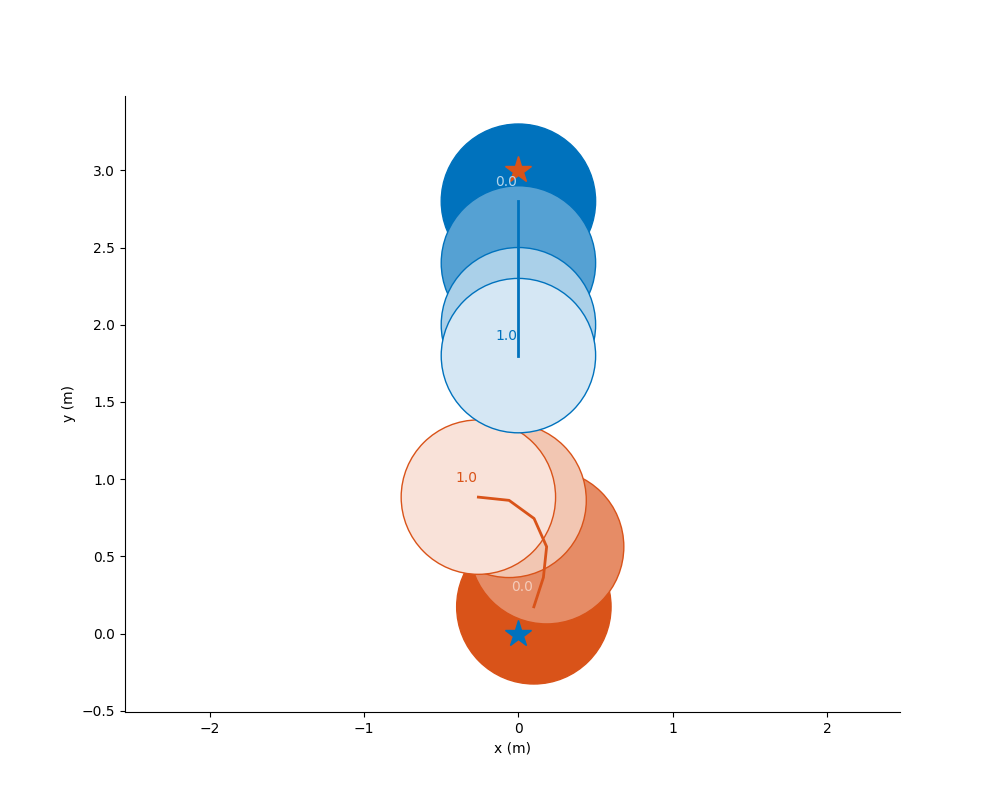}
	\vspace{-0.2in}
	\subcaption{}\label{fig:noise_crash_dqn}
	\end{minipage}%
	\begin{minipage}[t]{0.35\linewidth}
	\centering\includegraphics [trim=0 0 0 80, clip, width=1.0\textwidth, angle = 0]{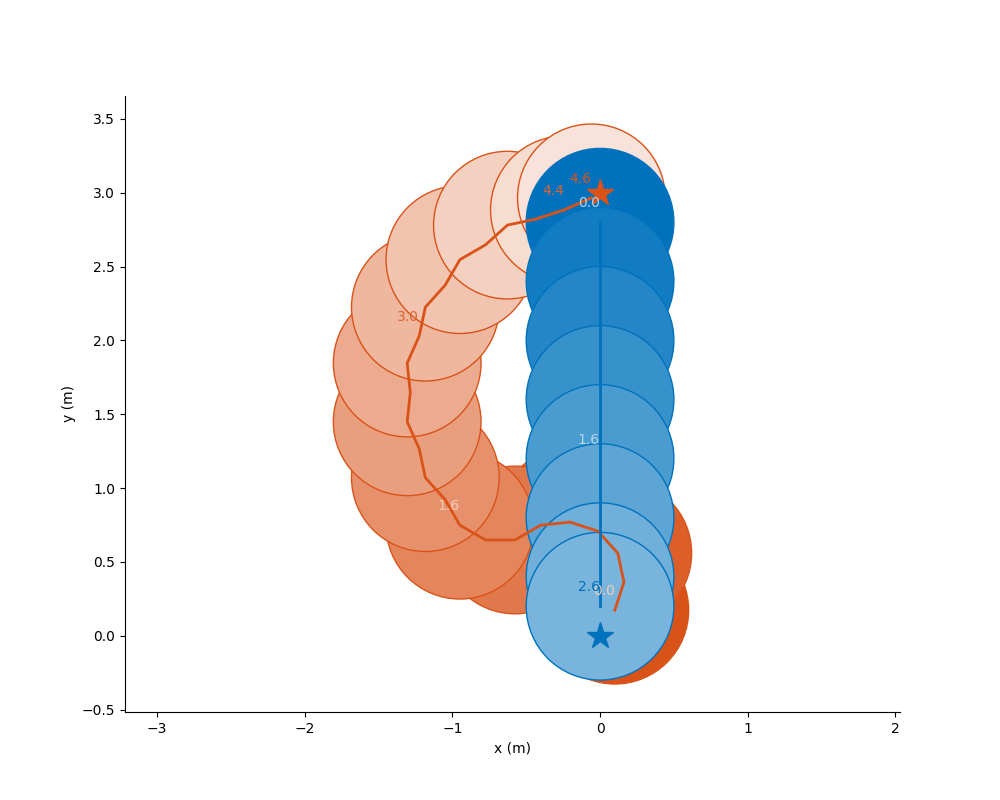}
	\vspace{-0.2in}
	\subcaption{}\label{fig:noise_crash_carrl}
	\end{minipage}
    \vspace{-.05in}
\caption[Collision avoidance through increased robustness]{DQN vs. CARRL. An agent (orange) tries to switch positions with a dynamic, non-cooperative obstacle (blue), while position observations are distorted with uniform noise within $\pm0.4m$ (each agent has radius $0.5m$). The nominal DQN policy, in~\cref{fig:noise_crash_dqn}, fails to avoid the obstacle due to the added noise. The CARRL policy ($\epsilon{=}0.1$), in~\cref{fig:noise_crash_carrl}, repeatedly considers the worst-case true obstacle state in its action selection, and successfully reaches the goal while avoiding a collision that would end the episode.}\label{fig:noise_crash}
\end{figure}

Visualization of the CARRL policy in particular scenarios offers additional intuition on the resulting policy.
In~\cref{fig:noise_crash}, an agent (orange) with radius $.5m$ (circle) observes a dynamic obstacle (blue) with added uniform noise $\sigma{=}.4m$ (not illustrated) on the position observation.
The nominal DQN agent, in~\cref{fig:noise_crash_dqn} is not robust to the noise and collides with the obstacle.
The CARRL policy, in~\cref{fig:noise_crash_carrl}, however, is robust to the small perturbation in the input space and successfully avoids the dynamic obstacle.
The resulting trajectories for several $\epsilon$ values are shown in~\cref{fig:eps_visual}, for the same uniform noise addition.
With increasing $\epsilon$ (toward right), the CARRL agent accounts for increasingly large worst-case position perturbations.
Accordingly, the agent avoids the obstacle increasingly conservatively, i.e., selects actions that leave more safety distance from the obstacle.

% At $\epsilon{=}1.0$, the states inside the $B_{\infty}(s_{adv}, \epsilon)$-ball are quite far from the training data, so the lower bound Q-values are inaccurate, leading to poor decisions.
%At $\epsilon{=}0.3$, the agent avoids the obstacle on the left instead of the right, since the Q-values for avoiding an obstacle on either side are approximately the same, and the action with the maximum Q-value might change under consideration of different input states in the $\epsilon$-ball
The non-dueling DQN used $2$, $64$-unit layers with hyperparameters: learning rate $2.05\texttt{e}{-4}$, $\epsilon$-greedy exploration frac. $0.497$, final $\epsilon$-greedy $0.054852$, buffer size $152\texttt{e}3$, $4\texttt{e}5$ training steps, and target network update frequency, $10\texttt{e}3$. The hyperparameters were found by running 100 iterations of Bayesian optimization with Gaussian Processes~\citep{Jasper_2012} on the maximization of the training reward.

\subsection{Generalization to the cartpole domain} %, Pong
Experiments in cartpole~\citep{barto1983neuronlike} show that the increased robustness to noise can generalize to another domain.
The reward is the time that a pole is successfully balanced (capped at $200$ steps).
The reward of a nominal DQN drops from $199$ to $141$ under uniform noise with $\sigma{=}.25$ added to all observations.
CARRL, however, considers the worst-case state, i.e., a state in which the pole is the closest to falling, resulting in a policy that recovers a reward of $180$ with $\epsilon{=}.05$, under the same noise conditions.
Hyperparameters for a $2$-layer, $4$-unit network were found via Bayesian Optimization. %The cartpole domain was evaluated with a $2$-layer, $4$-unit fully-connected network and hyperparameters found via Bayesian Optimization.

\iffalse
Real Advantage is that epsilon is unknown during test time
% TODO
Create robustness / noise plot with FGMT adversarial attacks
Visualize cartpole
Evaluate policy on Mountaincar, Inverted pendulum, Hopper, CHEETAH?, Maze, ..
3) Performance on real robot?!
Runtime of verification 

Even if our results degrade performance on standard data, this seems to inevitable as shown by~\citep{Tsipras_2019}
2.3) Table: Collisions, Goals, (Proxemix?), (Time to goal?)  

Additional speed-ups could be achieved by parallelizing the code and/or training the model towards ReLU stability, i.e. less ReLU switches~\citep{Xiao_2019}.
\fi

%!TEX root=main.tex

\section{Conclusion} \label{sec:conclusion}
This work adapted deep RL algorithms for application in safety-critical domains, by proposing an add-on certified defense to address existing failures under adversarially perturbed observations and sensor noise.
The proposed extension of robustness certification tools from computer vision into a deep RL formulation enabled efficient calculation of a lower bound on Q-values, given the observation uncertainty.
These guaranteed lower bounds were used to modify the action selection rule to provide maximum performance under worst-case observation perturbations.
The resulting policy (added onto trained DQN networks) was shown to improve robustness to adversaries and sensor noise, causing fewer collisions in a collision avoidance domain and higher reward in cartpole.
Future work will extend the guarantees to continuous action spaces and experiment with robotic hardware.

%===============================================================================

% The maximum paper length is 8 pages excluding references and acknowledgements, and 10 pages including references and acknowledgements

% The acknowledgments are automatically included only in the final version of the paper.
\clearpage
\acknowledgments{This work is supported by Ford Motor Company. The computation is supported by Amazon Web Services. The authors greatly thank Tsui-Wei (Lily) Weng for providing code for the Fast-Lin algorithm and insightful discussions.}

%===============================================================================

\setlength{\bibsep}{0pt plus 0.3ex}
%\bibitemsep=1in
\small{
\bibliography{biblio}  % .bib
}
\end{document}